\documentclass{IEEEcsmag}

\usepackage[colorlinks,urlcolor=blue,linkcolor=blue,citecolor=blue]{hyperref}

\usepackage{upmath}
\usepackage{cite}
\usepackage{amsmath,amssymb,amsfonts}
\usepackage{graphicx}
\usepackage{caption}
\usepackage{subcaption}
\usepackage{multirow}
\usepackage{amssymb}
\usepackage{amsmath}
\usepackage{soul}
\usepackage{booktabs}
\usepackage{algorithm}
\usepackage[noend]{algpseudocode}

\jvol{XX}
\jnum{XX}
\paper{8}
\jmonth{}
\jname{IEEE Intelligent Systems}
\pubyear{2021}

\begin{document}

\sptitle{\small{To appear in IEEE Intelligent Systems, 2021}}
\editor{}

\title{Dynamic Sampling and Selective Masking for Communication-Efficient Federated Learning}

\author{Shaoxiong Ji}
\affil{Department of Computer Science, Aalto University, Finland.}

\author{Wenqi Jiang}
\affil{Columbia University, USA.}

\author{Anwar Walid}
\affil{Nokia Bell Labs, USA.}

\author{Xue Li}
\affil{The University of Queensland, Australia.}

\begin{abstract}
Federated learning (FL) is a novel machine learning setting that enables on-device intelligence via decentralized training and federated optimization. Deep neural networks' rapid development facilitates the learning techniques for modeling complex problems and emerges into federated deep learning under the federated setting. However, the tremendous amount of model parameters burdens the communication network with a high load of transportation. This paper introduces two approaches for improving communication efficiency by dynamic sampling and top-$k$ selective masking. The former controls the fraction of selected client models dynamically, while the latter selects parameters with top-$k$ largest values of difference for federated updating. Experiments on convolutional image classification and recurrent language modeling are conducted on three public datasets to show our proposed methods' effectiveness.
\end{abstract}

\maketitle

\chapterinitial{The} widespread applications of mobile communication technology and personal mobile devices are turning machine learning to edge devices and making distributed agents more intelligent. 
Federated learning is a decentralized machine learning method, which uses distributed training on local users without sending the data to a central server~\cite{mcmahan2016communication}. 
Unlike centralized learning, FL trains the model without directly accessing private user data. Local computing and secure parameter transport can solve the security and privacy issues in traditional centralized training|~\cite{liu2020secure, zheng2020preserving}. 
In this new setting, users' data are stored securely on their own devices.

Federated deep learning takes deep neural networks as learning models on local devices and uses averaged model aggregation over sampled clients under the control of a central server. However, the humongous amount of transport cost in federated deep learning is one bottleneck for real-world applications because deep learning models have many parameters. Modern devices use wired communication or wireless communication with limited bandwidth. Thus, it is hard for current network transportation to handle such a large amount of transportation, making communication-efficient federated learning a critical mission.
 
There are two strategies to increase communication efficiency, i.e., to reduce the number of communication rounds between server and clients during federated training and to transport fewer parameters for each federated iteration.
To save the transportation cost, current work \cite{mcmahan2016communication} uses static sampling to select a fraction of client models, while other improvements use compression algorithms for more efficient communication.
This paper improves the vanilla federated averaging methods by dynamic sampling for client models and selective masking on client models' neural parameters. The former algorithm aims to reduce communication rounds, while the latter aims to sample a fraction of the parameters for transportation. Our proposed methods can also be combined with cutting-edge compression algorithms for furthering communication efficiency during model downloads and uploads. 

Our contributions are summarized in the following three folds.
\begin{itemize}
\item We propose a dynamic sampling strategy for federated averaging with exponential annealing of the sampling rate.
\item We propose selective masking on client models' neural parameters to save the amount of data during model transportation. 
\item Through experiments on convolutional image classification tasks on two popular image datasets and recurrent language modeling tasks on a typical text dataset, our proposed methods outperform their baseline methods in most cases. 
\end{itemize}

\section{Related~Work}
\label{sec:related}
Federated learning performs distributed training and learns efficiently from decentralized data using iterative averaging. 
It has wide range of application in wireless communication~\cite{chen2020joint,yang2020energy}.
Several improvements to the federated framework have been proposed, including per-user domain adaptation~\cite{peterson2019private} and attentive aggregation~\cite{ji2019learning, jiang2020decentralized}.
Communication efficiency is considered a vital evaluation metric of federated learning. To enable efficient communication, Kone\v{c}n\`{y} et al. \cite{konevcny2016federated} proposed sketched updating methods for reducing transport costs in federated learning and use random subsampling in their sketched updating method. Bonawitz et al. \cite{bonawitz2019towards} designed a scalable federated learning system architecture. 
Recently, several works proposed different strategies to improve the communication efficiency, including the sparse ternary compression~\cite{sattler2019robust}, Count Sketch-based compression~\cite{rothchild2020fetchsgd}, parameter quantization~\cite{reisizadeh2020fedpaq}, and federated ensembles~\cite{hamer2020fedboost}.

\section{Preliminaries}
\label{sec:pre}

\subsection{Federated~Averaging}

There is a central server for model aggregation in the federated setting and a set of client devices for local training. 
One typical algorithm is federated averaging, where the central server averages client models to obtain a global model that can well generalize distributed clients. For deep neural networks, the global model at $t$ is denoted as $\Theta_t = \{ W_t^1, \dots, W_t^i, \dots \}$ with multiple layers, where a real-value matrix $W_t^i \in \mathbb{R}^{d_1\times d_2}$ represents parameters in the $i$-th layer. The parameters of the global model are then distributed to clients via network transportation. For each selected client, the downloaded model is trained on its physical device using its data, with the trained local model in the $i$-th client at $t$-th communication round denoted as $\Theta_t^i$. Then, clients uploads trained models to the central server for model aggregation by computing the weighted average. The aggregated global model is computed in Eq. \ref{eq:avg}.
\begin{equation}
\label{eq:avg}
\Theta_{t+1}=\frac{1}{m}\sum_{i \in S}^m \Theta_t^i ,
\end{equation}
where $m$ is the total number of selected clients as a set $S$ in the distributed environment, 
For each layer in the neural network model, pair-wise matrix summation is calculated.
To consider data imbalance, federated averaging performs weighted averaging by taking number of training samples as weights. The aggregation of client models is calculated as
\begin{equation}
\Theta_{t+1}=\frac{1}{m} \sum_{i \in S}^{m} \frac{n_i}{n}\Theta_{t}^{i},
\end{equation}
where $n_i$ is the number of training instances in the $i$-th device and $n=\sum_{i=0}^m n_i$ is the total number of training samples of all selected clients.

\subsection{Static~Sampling}
Federated learning applies static sampling by selecting a random fraction of clients for federated averaging. The server initially sets a sampling rate of $C$, and then it waits for updates from clients. Once there are enough updates to meet the sampling rate, the server will stop receiving updates and turn to federated averaging. During this procedure, the sampling rate of $C$ remains unchanged, which is why it is a so-called static sampling. 

\subsection{Random~Masking}
In a distributed computing environment, the bottleneck of federated learning is the vast amount of transport cost. The bandwidth of the central server is fixed when receiving updates from distributed clients. 
A straightforward strategy is to randomly select a part of updated model parameters for transportation to save the transportation cost. We term this as random masking because a random mask is applied to the parameters of each neural layer. 
Random masking is randomized by $randi$ function given a random seed in each client, which generates a matrix of $A\in \mathbb{R}^{a\times b}$ with $\gamma$ of ones. A certain proportion of parameters is masked via the pair-wise product of $A$ matrix and parameters in each layer. Then, the masked model is compressed when uploaded to the central server. 

\section{Proposed~Methods}
\label{sec:methods}
We propose two strategies to save the communication cost. They are dynamic sampling and selective masking. The iterative model aggregation method uses federated averaging \cite{mcmahan2016communication}. Dynamic sampling is proposed to control the sampling rate of client models. At the same time, selective masking chooses parameters according to the absolute difference value of parameters with a preset proportion of $\gamma$ and saves communication costs.

\subsection{Dynamic~Sampling}
Static subsampling uses a fixed subsample rate throughout the training process, no matter the epochs and training steps. This method is secure for implementation by evenly choosing the number of clients to enable model aggregation. We propose dynamic sampling with a high sampling rate first and then decrease the sampling rate during each communication. Our motivation is to accelerate convergence at the beginning of federated learning by involving more clients for model aggregation at the very beginning. Once a more generalized federated model is trained based on the initialization, our method dynamically decreases the number of clients for model aggregation to save communication costs. Even though it costs more at the beginning of federated training, its selected number of clients model declines rapidly after several rounds of training. The declining rate of sampling rate can be chosen accordingly to ensure that the total amount of parameter transportation of dynamic sampling is fewer than its counterpart of static sampling after individual rounds of communication.

Our proposed dynamic subsampling method uses an exponential decay rate to anneal the sampling rate in the training process, where the subsample rate $R$ is a function of current epoch $t$ and decay coefficient $\beta$ as shown in Eq. \ref{eq:decay}.
\begin{equation}
\label{eq:decay}
R(t, \beta)=\frac{1}{\exp({\beta t})}
\end{equation}
With the decreasing rate multiplied with a preset initial sample rate $C$, we get the dynamic sampling rate as $c=\frac{C}{\exp(\beta t)}$ at $t$-th training round.
With the increase of communication rounds, the sampling rate becomes very small, making less than one client selected for model aggregation. In practice, the minimum number of selected client models is set to two. 
Integrated with federated averaging, the algorithm of the dynamic sampling is written in Algorithm \ref{alg:dynamic}. The core difference between static sampling is the dynamically changed sampling rate. 

\begin{algorithm}[ht]
\small
\caption{Federated~Averaging~with~Dynamic~Sampling}
\label{alg:dynamic}
\begin{algorithmic}[1]
\State $C \in \mathbb{R}$ is a constant of initial sampling rate; $c \in \mathbb{R}$ is a variable of the sampling fraction of clients; $R \in \mathbb{N}^+$ is the value of preset communication rounds; $k \in \mathbb{R}$ is a constant of the decay coefficient in Eq. \ref{eq:decay}. 
\State \textbf{Input}: a set of $M$ registered clients $S=\{s_1, \dots, s_M\}$
\State \textbf{Output}: updated global parameters $\Theta_{t+1}$
\Procedure{Federated~Averaging}{$S$} \Comment{Run on a central server}
\State Initialize model $\Theta_0$, set the decay efficiency of sampling $\beta$.
\For{$t~=~1:R$}
	\State Initialize an empty list $L$
	\State Calculate sample rate $c=\frac{C}{\exp(\beta t)}$
	\State Number of sampled clients $m=\max(c*M, 1)$
	\While{len(L) $<$ m}	\Comment{Dynamic~sampling}
		\State Send connection request to clients
		\If{ACK from $i$-th client}
			\State call ClientUpdate($i, \Theta_t$)
			\State $\Theta_t^k \gets~receiveParam(i)$
			\State $L.add(\Theta_t^i)$
		\EndIf
	\EndWhile
	\State $\Theta_{t+1}=\frac{1}{m}\sum_{i \in C}^m \frac{n_i}{n} \Theta_t^i $
\EndFor
\EndProcedure
\end{algorithmic}
\end{algorithm}

\subsection{Selective~Masking}
This subsection turns to another improvement - the selective masking technique. Model parameters stored in client devices take up most of the transportation cost. When transporting data from user devices to central servers, the random masking algorithm randomly selects some computed updates and discards the rest. However, this method is less heuristic because it is unable to select prior updates. Consequently, some important updates can be discarded when randomly select parameters.

We propose selective masking to consider the importance of model parameters in each local training. Given a static masking rate on the proportion of model parameters as the selective criteria, only model parameters with the largest absolute difference are selected proportionally for federated aggregation and model updating. 
First, the difference of current model parameters of $i$-th layer and updated counterpart in the next time step is calculated as
\begin{equation}
D_t^i = \mid W_{t+1}^{i}-W_{t}^{i} \mid
\end{equation}
Given the masking proportion of $\gamma$, top-$k$ largest values are selected together with their indices, where $k$ equals $\gamma$ multiplied with the number of elements of the weight matrix. A mask matrix is generated with the selected indices as $M$, which contains $1-\gamma$ zeros and $\gamma$ ones. Then, the masked weight matrix is calculated by the pair-wise product of mask matrix and full weight matrix as
\begin{equation}
W_{t+1}^{i}=M \otimes W_{t+1}^{i}.
\end{equation}
Finally, the masked parameters are compressed and ready for transportation via client-server communication.

The algorithm of on-device training with selective masking is written in Algorithm \ref{alg:selective}, where the $\operatorname{genMask}$ function generates the mask matrix according to the top-$k$ indices.

\begin{algorithm}[ht]
\small
\caption{On-Device~Training~with~Selective~Masking}
\label{alg:selective}
\begin{algorithmic}[1]
\State $\gamma$ is the proportion of masked parameters; B is the local mini-batch size; E is the number of local epochs; $\eta$ is the learning rate.
\State \textbf{Input}: ordinal of user $k$, user data $X$.
\State \textbf{Output}: updated user parameters $\Theta_{t+1}$ at $t+1$.
\Procedure{Client Update}{$k$,~$\Theta$} \Comment{Run on the $k$-th client}
\State $\textrm{B} \gets (split~user~data~X~into~batches)$
\For{each~local~epoch~i~from~1~to~E}
	\For{batch~b~$\in~\textrm{B}$}
		\State $ \Theta_{t+1} \gets \Theta_t - \eta \nabla\textit{L}(\Theta_t) $
	\EndFor
\EndFor
\For{each~layer~i~from~1~to~l} 
	\State $(a,~b)=shape(W_{t+1}^i)$
	\State $D = \mid W_{t+1}^i - W_{t}^i \mid$
	\State $indices, values = \operatorname{topk}(D,~\gamma)$
	\State $M=genMask([0~1],~a,~b,~indices)$ 
	\State $W_{t+1}^i=M\otimes W_{t+1}^i$ 
\EndFor
\State send~$\Theta_{t+1}$~to~server 
\EndProcedure
\end{algorithmic}
\end{algorithm}

\section{Experiments}
\label{sec:exp}

\subsection{Datasets~and~Settings}
\subsubsection{Datasets} 
We perform several neural architectures on two typical tasks - image classification and language modeling - on two image datasets of MNIST  and CIFAR-10 and a natural language dataset of WikiText-2. They are typical public datasets for training machine learning algorithms. 
MNIST\footnote{Available at \url{http://yann.lecun.com/exdb/mnist/}} and CIFAR-10\footnote{\url{https://www.cs.toronto.edu/~kriz/cifar.html}} are two widely used datasets with hand-written digits and objects respectively.
WikiText-2\footnote{\url{https://s3.amazonaws.com/research.metamind.io/wikitext/wikitext-2-v1.zip}} is a dataset for word-level language modeling, which contains more than 2 million tokens extracted from Wikipedia. Our experiments are performed on these three widely-used datasets for validation. Statistics of these three datasets are summarized as in Table \ref{tab:stat}.
\begin{table}[htp]
\scriptsize
\caption{A summary of datasets}
\begin{center}
\begin{tabular}{|c|c|c|c|}
\toprule
Dataset & Type & \# train & \# test \\
\midrule
MNIST & image & 60,000 & 10,000\\
CIFAR-10 & image & 50,000 & 10,000\\
Wikitext-2 & token & 2,088,628 & 245,569\\
\bottomrule 
\end{tabular}
\end{center}
\label{tab:stat}
\end{table}%

\subsubsection{Data~Partitioning}
Those three datasets are designed for training a centralized machine learning model. Data partitioning is adopted for generating decentralized datasets by sampling the whole dataset under independent and identical distribution (IID). We followed the data partitioning rule of MNIST and CIFAR-10 proposed by McMahan et al. \cite{mcmahan2016communication}. Each partitioned subset of the whole dataset is regarded as a private dataset in a client device. The same partitioning rule is then applied to WikiText-2 to construct the IID federated dataset. 

\subsubsection{Settings}
We conduct both convolutional and recurrent neural networks according to specific tasks. For the image classification task, we use LeNet as a client model, and for furthering the effect of a large-scale model, VGG is implemented as a client model on the CIFAR-10 dataset. The language modeling task deals with sequential text data. Thus, we use long short-term memory networks (LSTM) to capture sequential dependency with tied and untied word embedding. All neural models are implemented by the PyTorch framework and accelerated by Nvidia K40m GPU. 

Our aim is to enable communication-efficient federated learning. Thus, transportation cost is considered for evaluation, which is related to sampling rate, masking rate and communication round. Taking a single communication between one client and the server with full model parameters as unit, it is calculated as 
\begin{equation}
f(\beta, \gamma) = \frac{\gamma}{R} \sum_{t=1}^R \frac{C}{\exp({\beta t})} 
\end{equation}
for $R$ rounds of client-server communication. 
In the following section, we take transportation costs together with prediction accuracy as evaluation metrics. 

\subsection{Convolutional~Image~Classification}
We first begin with the task of convolutional image classification in the federated setting. 
Two groups of comparisons are conducted on static versus dynamic sampling and random versus selective masking on MNIST digits classification using LeNet. Then, an analogical comparison is further perform using a large-scale VGG model on CIFAR-10 objects classification.
Due to the sample rate decay, the dynamic sampling method can train more federated rounds than the static method, given the same transport cost and the same initial sampling rate. For example, with a decay coefficient of 0.1 and the same amount of transportation cost, the dynamic method can update 31 epochs. In comparison, the static method can only train ten epochs of updates. In the following experimental analysis, a convolutional image classification task is performed under separate comparison first and then with two methods combined for federated training. 

\subsubsection{Static Versus Dynamic Sampling}
\begin{figure}[htbp]
\begin{center}
\begin{subfigure}[b]{0.22\textwidth}
	\includegraphics[width=\textwidth]{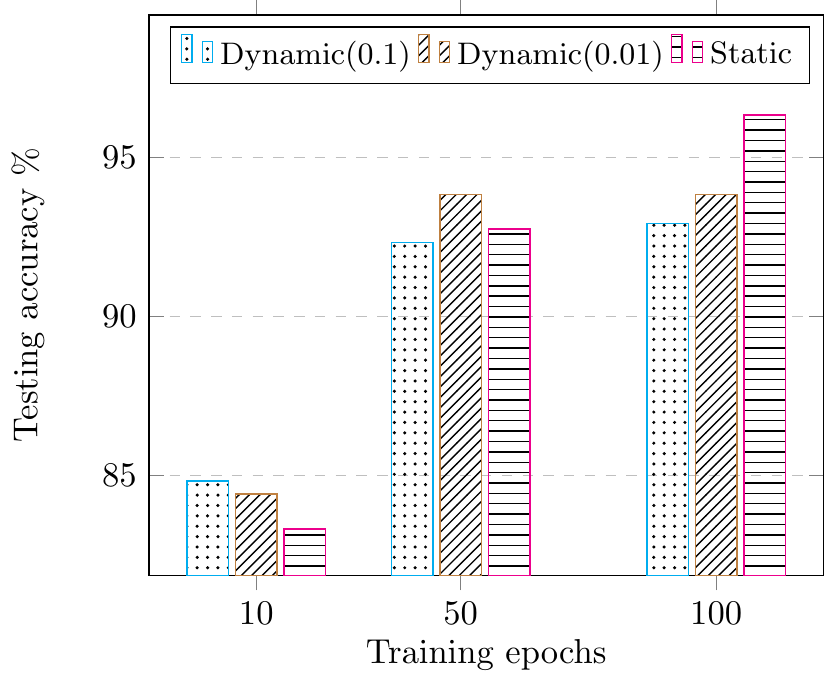}
	\caption{Prediction accuracy}
	\label{figs:mnist-sampling}
\end{subfigure}
\quad
\begin{subfigure}[b]{0.22\textwidth}
	\includegraphics[width=\textwidth]{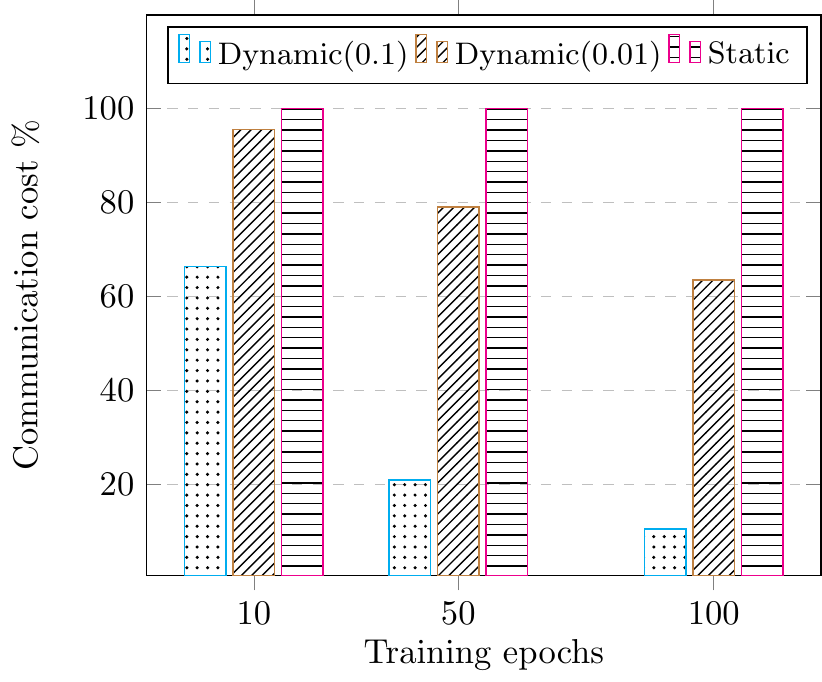}
	\caption{Communication cost}
	\label{figs:mnist-cost}
\end{subfigure}
\caption{Static versus dynamic sampling with 100\% clients for initial model aggregation on MNIST dataset. For dynamic sampling, decay coefficient is set as 0.01 and 0.1.}
\label{figs:mnist}
\end{center}
\end{figure}

Sampling strategies are conducted in static and dynamic manners. We take the whole cohort of clients as initial training. The static sampling method keeps this rate during the whole process of federated training. At the same time, the dynamic counterpart takes an exponential decay on the sampling rate with the decay coefficient in Eq. \ref{eq:decay} of 0.01 and 0.1. Results of prediction accuracy and communication cost are reported in Fig. \ref{figs:mnist} after 10, 50, and 100 rounds of federated training. According to Fig. \ref{figs:mnist-sampling}, with the increase of training epochs, prediction accuracy grows steadily for all three settings. For fewer federated training rounds (that is, 10 and 50 rounds), the testing accuracy of static sampling is more insufficient than dynamic sampling with 0.01 as the decay coefficient. However, when the sampling rate drops faster (i.e., a higher decay coefficient of 0.1), testing accuracy of dynamic sampling is imparted. After 50 and 100 rounds of training, static sampling gains a better prediction performance than the dynamic sampling method. 
As for the transportation cost during federated communication, the static method takes 100\% of transportation, while the dynamic sampling method can efficiently save communication costs. With the increase of training epochs and the decay coefficient, many more rounds of client-server communication have been saved, as illustrated in Fig. \ref{figs:mnist-cost}.

\subsubsection{Random Versus Selective Masking}
\begin{figure}[htbp]
\begin{center}
\includegraphics[width=0.3\textwidth]{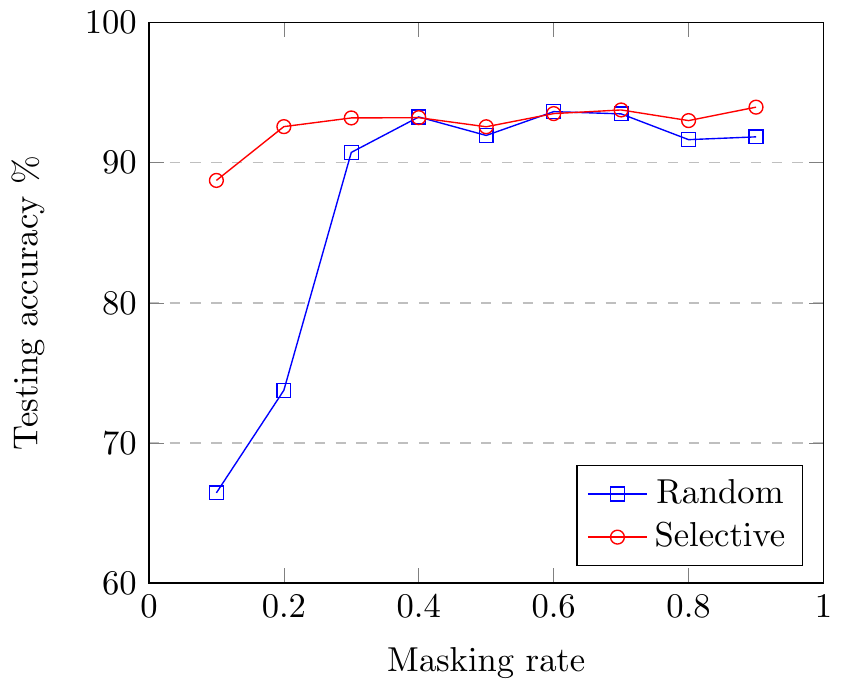}
\caption{Random masking versus selective masking with static sampling rate of 0.1 for 10 rounds federated training on MNIST dataset.}
\label{figs:cifar}
\end{center}
\end{figure}

After that, we evaluate the performance of the selective method using top-$k$ masking. This section fixes the sampling rate to be 0.1 and conducts experiments with random masking and selective masking. For a fair comparison, these two methods use the same hyperparameter setting for ten rounds of training, and the learning rate is set to 0.01. Experimental results are reported in Fig. \ref{figs:cifar} where the masking rate varies from 0.1 to 0.9. With a relatively higher masking rate, the testing accuracy of random masking and selective masking is close. Selective masking performs a bit higher testing accuracy for a masking rate of 0.8 and 0.9. When a large number of parameters are discarded with a rate of 0.1 and 0.2, the performance of random masking drops dramatically. According to this result, our proposed top-$k$ selective masking method can maintain a stable performance to save communication costs even with a high proportion of parameters ignored. 

\subsubsection{Combined~Experiment}
\begin{figure}[htbp]
\begin{center}
\begin{subfigure}[b]{0.22\textwidth}
	\includegraphics[width=\textwidth]{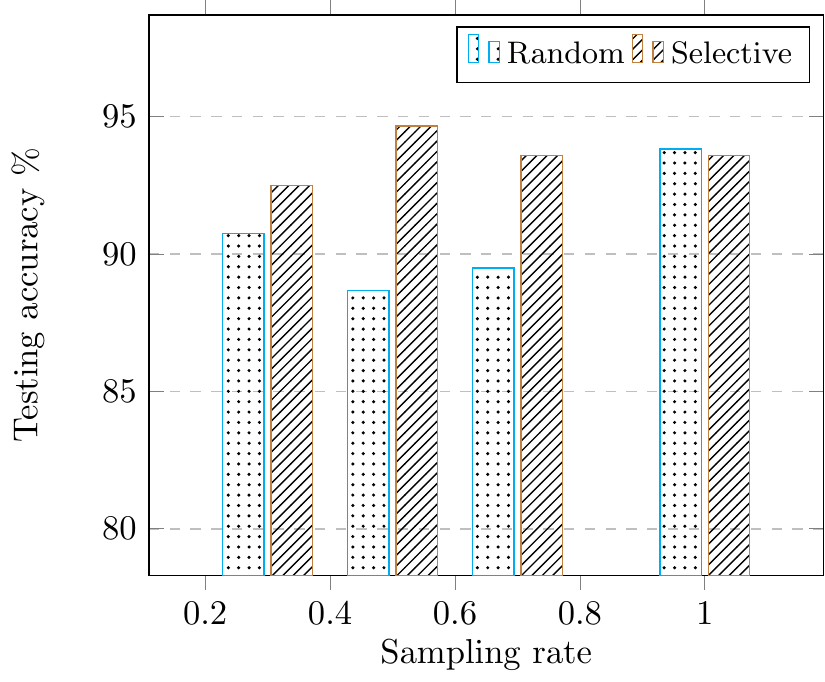}
	\caption{Decay coefficient = 0.01}
	\label{figs:mnist-sampling-decay1}
\end{subfigure}
\quad
\begin{subfigure}[b]{0.22\textwidth}
	\includegraphics[width=\textwidth]{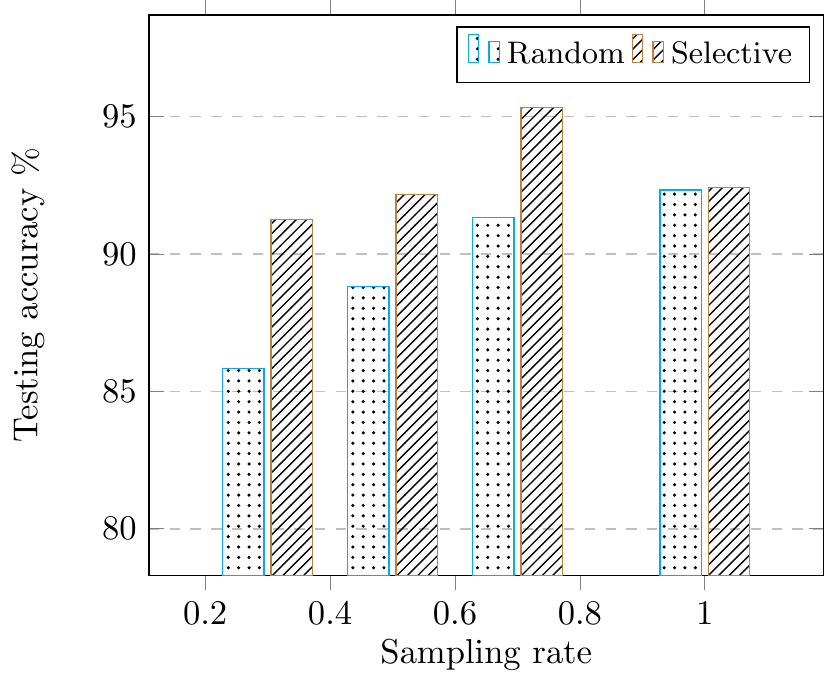}
	\caption{Decay coefficient = 0.1}
	\label{figs:mnist-sampling-decay2}
\end{subfigure}
\caption{Random masking verses selective masking with dynamic sampling using sample coefficient of 0.01 and 0.1 after 50 training epochs on MNIST dataset.}
\label{figs:mnist-sampling-masking}
\end{center}
\end{figure}

After an independent comparison of two proposed methods, we combine them into federated training for evaluation. In this section, four initial sampling rates of 0.3, 0.5, 0.7, and 1.0 are included in the dynamic sampling method. As for the selective masking, two decay coefficients of 0.01 and 0.1 are used. Experimental results after 50 training rounds are shown in the two bar charts of Fig. \ref{figs:mnist-sampling-masking}. Selective masking outperforms random masking in the dynamic sampling setting of these two cases except when the initial sampling rate equals 1 with the decay coefficient of 0.01. 

\subsubsection{Experiments~on~CIFAR-10}
\begin{figure}[htbp]
\begin{center}
\includegraphics[width=0.3\textwidth]{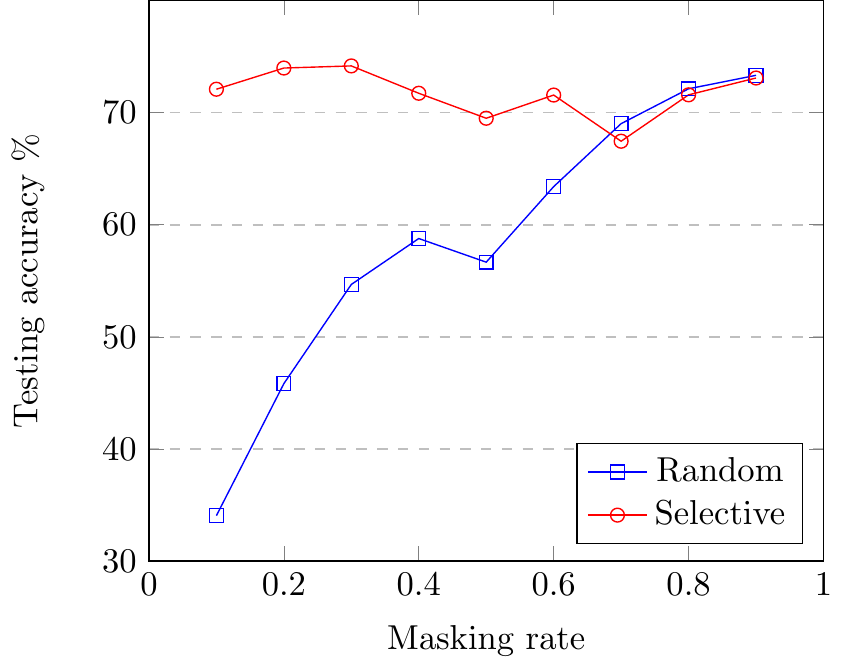}
\caption{Aggregated prediction accuracy using random masking and selective masking with VGG model on CIFAR-10 after 100 federated training rounds.}
\label{figs:cifar-masking}
\end{center}
\end{figure}

Then, we further experiment on the CIFAR-10 dataset using the VGG-16 model to evaluate the performance on a large-scale model. The aggregated prediction accuracy of random and selective masking after 100 federated training rounds is shown in Fig. \ref{figs:cifar-masking} where static sampling with a 100\% sampling rate is applied. 
Notice that we aim to compare the performance of masking methods but not the image classification model's performance. Thus, we do not tune client learners hard. All the experiments have not achieved the state-of-the-art results of centralized training as the comparison is conducted in the federated setting with limited communication rounds. However, our comparison is reported with the same set of client learners to ensure a fair comparison of the two proposed methods and their counterparts. As we can see from that table, the top-$k$ selective masking method outperforms random masking for a masking rate from 0.1 to 0.6. When the masking rate is high, these two methods gain similar testing accuracy. Our proposed top-$k$ selective masking can maintain satisfactory performance with a large proportion of parameters saved in federated training. 
 
\begin{figure}[htbp]
\begin{center}
\begin{subfigure}{0.22\textwidth}
	\includegraphics[width=\textwidth]{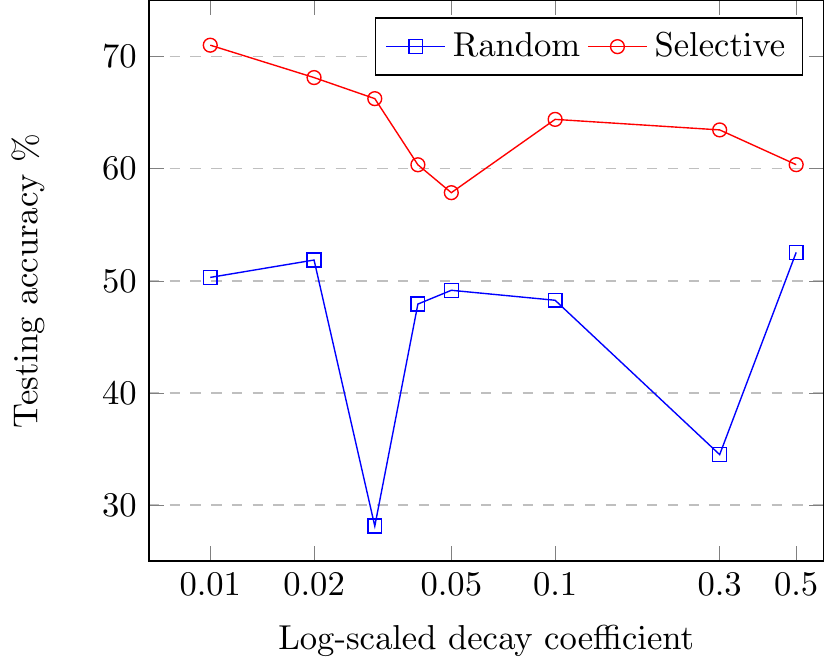}
	\caption{Masking rate $=$ 0.3}
	\label{figs:cifar-masking3}
\end{subfigure}
\quad
\begin{subfigure}{0.22\textwidth}
	\includegraphics[width=\textwidth]{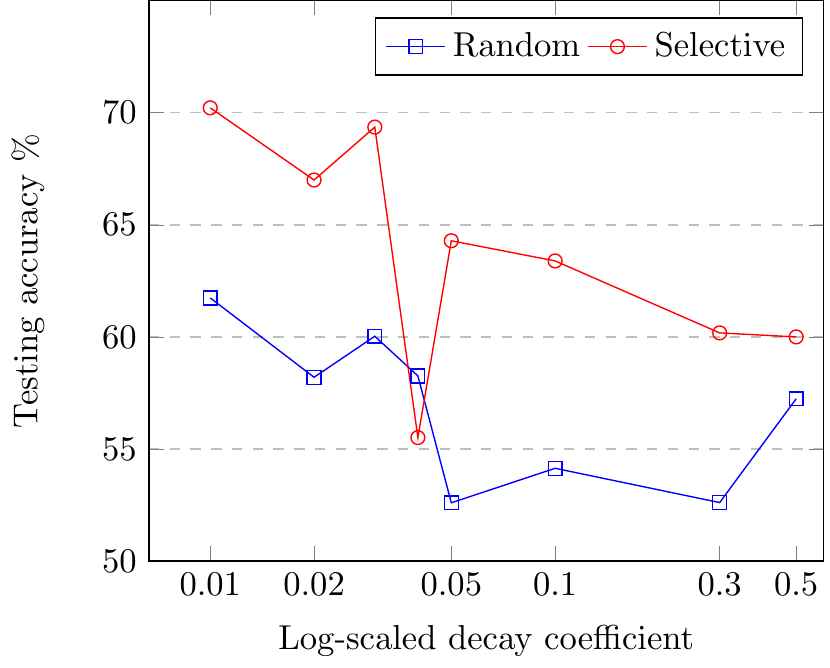}
	\caption{Masking rate $=$ 0.5}
	\label{figs:cifar-masking5}
\end{subfigure}
\quad
\begin{subfigure}{0.22\textwidth}
	\includegraphics[width=\textwidth]{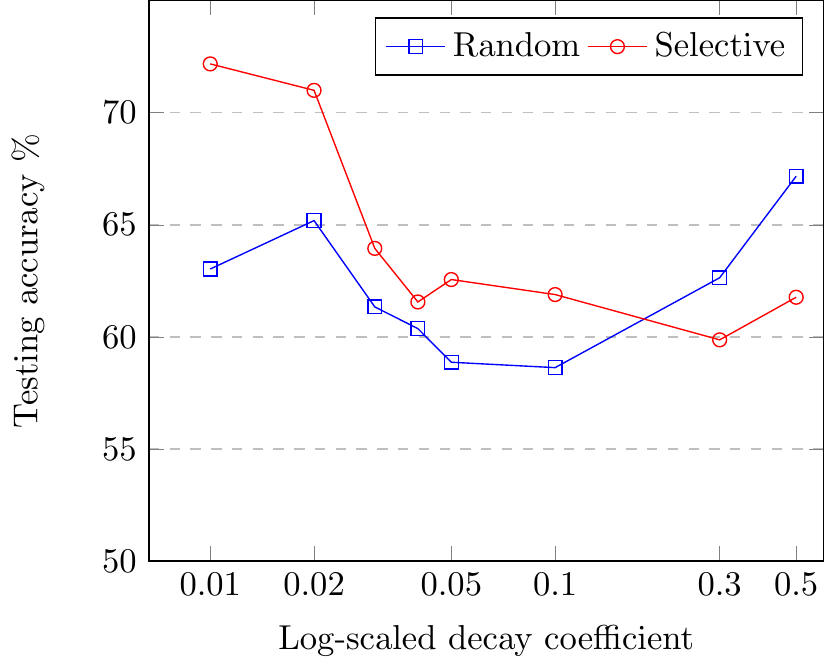}
	\caption{Masking rate $=$ 0.7}
	\label{figs:cifar-masking7}
\end{subfigure}
\quad
\begin{subfigure}{0.22\textwidth}
	\includegraphics[width=\textwidth]{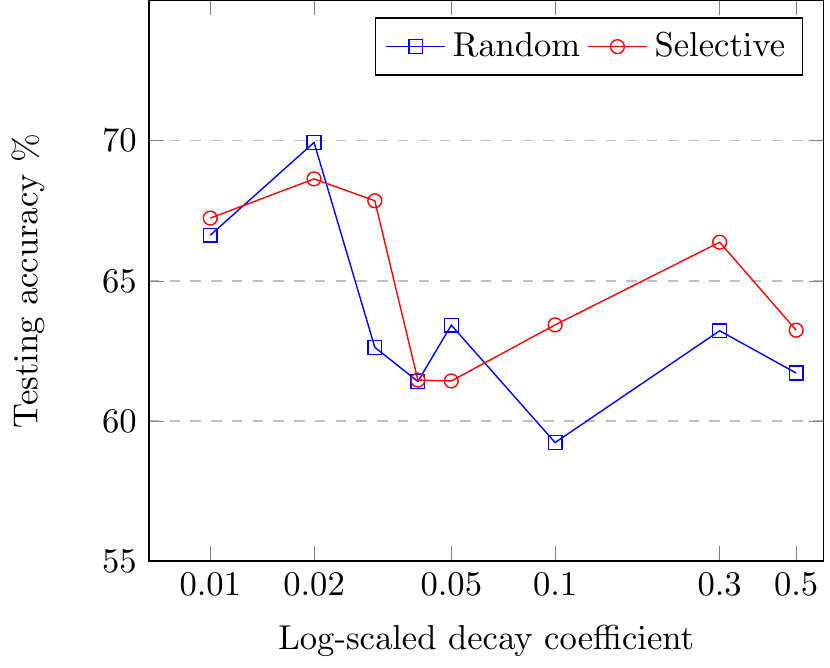}
	\caption{Masking rate $=$ 0.9}
	\label{figs:cifar-masking9}
\end{subfigure}
\caption{The effect of different decay coefficients on dynamic sampling with federated aggregation using different masking rates on CIFAR-10.}
\label{figs:cifar-decay}
\end{center}
\end{figure}

Besides, we conduct experiments with both sampling and masking strategies adopted for deeper analysis. This comparison aims to evaluate the effect of the decay coefficient in dynamic sampling with masked updating applied. The results of using masking rates of 0.3, 0.5, 0.7, and 0.9 are reported in Fig. \ref{figs:cifar-decay} where the $x$-axis is log-scaled. These figures show that when the masking rate is 0.3, selective masking outperforms random masking with all settings of the decay coefficient. For the masking rate being 0.5 and 0.7, selective masking gains better testing accuracy in most cases. With a higher masking rate of 0.9, the performance gap between these two methods is narrow. Generally, with a larger decay coefficient (more communication-efficient), the performance experiences a fluctuation and decreases to a relatively low level when the decay coefficient is set as 0.5. 

\subsection{Recurrent~Language~Modeling}
Mobile keyboard suggestion with private neural language modeling is a typical application of federated learning, which interacts with users to provide supervised labels. 
This section models the next word prediction in the mobile keyboard as private RNN-based language modeling. Specifically, we adopted the gated recurrent unit (GRU) as the client learner. GRU is a simplified variant of the long short-term memory (LSTM) network, which suits saving communication costs with fewer parameters. The natural language corpus usually has a vast vocabulary. To further communication-efficient federated learning, tying word embedding and word classifier are introduced by using shared parameters. In this section's experiments, tried embedding is applied. For the evaluation metric, we use the aggregated perplexity. Perplexity is a standard metric for language modeling tasks. According to its definition, lower perplexity means better performance. 

\begin{figure}[htbp]
\begin{center}
\includegraphics[width=0.45\textwidth]{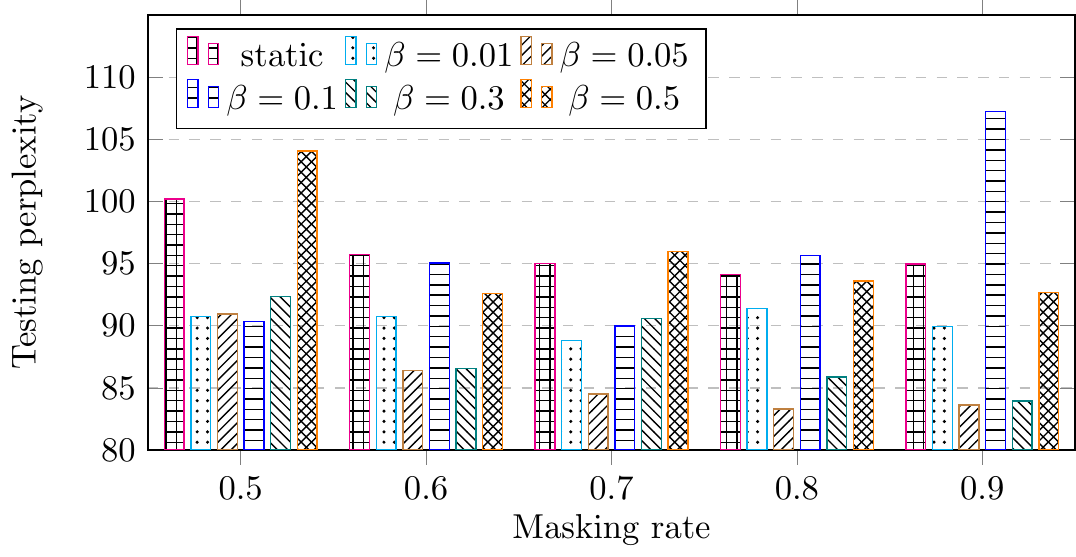}
\caption{Static versus dynamic sampling with masked updating using different masking rate after 50 communication rounds on WikiText-2.}
\label{figs:wikitext-sampling}
\end{center}
\end{figure}

We first compared the effect of sampling strategies. Fig. \ref{figs:wikitext-sampling} takes 50 rounds of client-server communication with different rates of masking to compare the performance of static and dynamic sampling. This bar chart shows that dynamic sampling achieves a lower perplexity in most cases, excluding $\beta=0.5$ with a masking rate of 0.5 and 0.7 and $\beta=0.1$ with a masking rate of 0.8 and 0.9.

\begin{figure}[htbp]
\begin{center}
\includegraphics[width=0.3\textwidth]{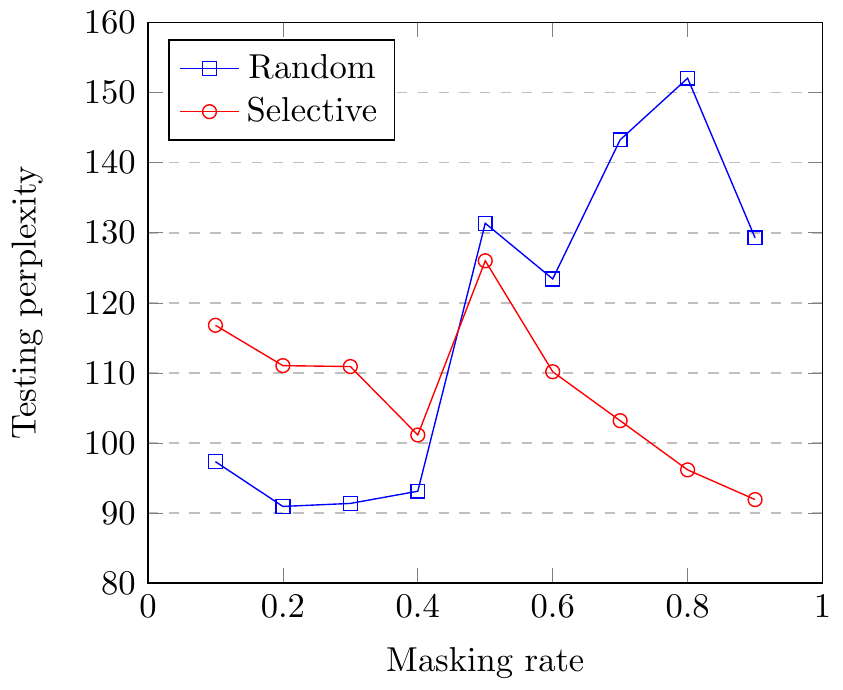}
\caption{Random versus selective masking with different masking rates on WikiText-2.}
\label{figs:wikitext-masking}
\end{center}
\end{figure}

Secondly, we compared random masking with selective masking. Results using different masking rates are reported in Fig. \ref{figs:wikitext-masking}. Our proposed selective masking is better for larger masking rates. Surprisingly, random masking gains better performance when the masking rate is low. It is hard to interpret why random masking is better, even when many parameters are discarded for updating. One possible guess is that the randomness improves the generalization of aggregated recurrent model, making the testing perplexity decrease. 

\subsection{Discussion}
Comprehensive experiments on a single computing node are conducted to mimic the federated setting in this section. Two tasks of convolutional image classification and recurrent language modeling are performed with comparative analysis. Our proposed achieves competitive performance in most experimental settings, which provides empirical approaches for saving communication costs in the federated setting.  
For simplicity, we ignore the loss of network transmission. However, admittedly, the federated setting in the real-world environment is more complicated, which requires further simulation experiments. 
Deep neural models have a large number of parameters. Taking VGG-16 as an example, its total number of parameters is more than one hundred million. High-speed network techniques are also required. Due to the lack of computing resources, we leave experimental simulations on multiple computing nodes for future work.

\section{Conclusion}
\label{sec:conclusion}
Federated learning decouples modeling training and data accessing, which protects data privacy. However, it incurs a communication cost issue when combines with large-scale deep neural networks. This paper proposes two empirical approaches - dynamic sampling and selective masking to save communication costs while ensuring satisfying prediction performance. The proposed two strategies can save server-client communication and save the number of model parameters for each transmission. Experiments on convolutional image classification and recurrent language modeling show that our proposed methods gain competitive results. 

\bibliographystyle{plain}
\bibliography{ref-fed-dyn}

\end{document}